%% file: main.tex
\documentclass[letterpaper]{article}
\usepackage{aaai25}
\input{preamble.tex}
\nocopyright

\title{Deep Learning for Generalised Planning with Background Knowledge}

\author{Dillon Z. Chen$^{1,2}$, Rostislav Hor{\v c}{\'\i}k$^{3}$, Gustav {\v S}{\'\i}r$^{3}$}
\affiliations{
    $^1$LAAS-CNRS, University of Toulouse \\
    $^2$The Australian National University \\
    $^3$Czech Technical University in Prague \\
    dchen@laas.fr, xhorcik@fel.cvut.cz, gustav.sir@cvut.cz
}

\begin{document}

\maketitle

\begin{abstract}
Automated planning is a form of declarative problem solving which has recently drawn attention from the machine learning (ML) community. ML has been applied to planning either as a way to test `reasoning capabilities' of architectures, or more pragmatically in an attempt to scale up solvers with learned domain knowledge. In practice, planning problems are easy to solve but hard to optimise. However, ML approaches still struggle to solve many problems that are often easy for both humans and classical planners. In this paper, we thus propose a new ML approach that allows users to specify background knowledge (BK) through Datalog rules to guide both the learning and planning processes in an integrated fashion. By incorporating BK, our approach bypasses the need to relearn how to solve problems from scratch and instead focuses the learning on plan quality optimisation. Experiments with BK demonstrate that our method successfully scales and learns to plan efficiently with high quality solutions from small training data generated in under 5 seconds.
\end{abstract}

\input{sections/introduction.tex}

\input{sections/background.tex}

\input{sections/bk-domains.tex}

\input{sections/lrnn.tex}

\input{sections/experiments.tex}

\input{sections/related-work.tex}

\section{Conclusion and Future Work}\label{sec:conclusion}
We proposed a new learning for planning paradigm aimed at improving solution quality in planning domains that are easy to solve but hard to optimise.
Our approach employs ``background knowledge'' in the form of declarative Datalog rules, representing a satisficing strategy for a planning domain, along with a general message passing scheme.
These rules are then parameterised and trained from data in an end-to-end differentiable manner, resulting in plan quality improvements over satisficing policies in the experiments.
Our new approach opens up several avenues of future work for computing high quality plans more efficiently, such as by incorporating our generalised policies into new anytime planning and heuristic or local search algorithms.

\section{Acknowledgements}
The authors would like to thank Sylvie Thi\'ebaux and Marcel Steinmetz for discussions about the work.
The computing resources for the project was supported by the Australian Government through the National Computational Infrastructure (NCI) under the ANU Startup Scheme.
This work has received funding from the European Union's Horizon Europe Research and Innovation program under the grant agreement TUPLES No. 101070149.

\bibliography{aaai25}

\end{document}

%% file: preamble.tex
\usepackage{times}  
\usepackage{helvet}  
\usepackage{courier}  
\usepackage[hyphens]{url}  
\usepackage{graphicx} 
\usepackage{natbib}  
\usepackage{caption} 
\usepackage{bm}         
\usepackage[utf8]{inputenc}
\usepackage{lipsum}
\usepackage{amssymb,amsthm,amsmath}
\usepackage{enumitem}
\usepackage{caption}
\usepackage{subcaption}
\usepackage{centernot}
\usepackage{complexity}
\usepackage[ruled,noline,linesnumbered,noend]{algorithm2e} 
\usepackage{mathrsfs}   
\usepackage{tabularx}   
\usepackage{multirow}   
\usepackage{array}      
\usepackage{booktabs}   
\usepackage{colortbl}   
\usepackage{xargs}      
\usepackage{tikz}
\usepackage{tikz-cd}
\usepackage{marginnote} 
\usepackage{afterpage}
\usepackage{pdflscape}
\usepackage{environ}    
\usepackage{braket}

\usetikzlibrary{positioning, fit, calc, shapes, arrows}

\DeclareMathOperator*{\pre}{pre}
\DeclareMathOperator*{\add}{add}
\DeclareMathOperator*{\del}{del}

\newclass{\N}{N}
\newclass{\CountingLogic}{C}
\newclass{\coNTIME}{coNTIME}
\newclass{\coNSPACE}{coNSPACE}
\newclass{\coNPSPACE}{coNPSPACE}
\newclass{\EXPTIME}{EXPTIME}
\newclass{\NEXPTIME}{NEXPTIME}
\newclass{\coNEXPTIME}{coNEXPTIME}
\newclass{\NEXPSPACE}{NEXPSPACE}
\newclass{\coNEXPSPACE}{coNEXPSPACE}
\newclass{\ASPACE}{ASPACE}
\newclass{\ATIME}{ATIME}
\newclass{\APSPACE}{APSPACE}
\newclass{\AEXPTIME}{AEXPTIME}
\newclass{\AEXPSPACE}{AEXPSPACE}

\newtheorem{property}{Property}
\newtheorem{theorem}{Theorem}[section]

\theoremstyle{definition}
\newtheorem{example}[theorem]{Example}
\theoremstyle{definition}
\newtheorem{definition}[theorem]{Definition}

\def\N{\mathbb{N}}

\def\a{\alpha}

\renewcommand{\phi}{\varphi}

\def\la{\leftarrow}

\newcommand{\gen}[1]{\left< #1 \right>}

\newcommand{\sett}[1]{\left\{ #1 \right\}}

\newcommand{\msetbig}[1]{\Bigl\{ \!\! \Bigl\{ #1 \Bigr\} \!\! \Bigr\}}

\newcolumntype{Y}{>{\raggedleft\arraybackslash}X}

\newcommand{\zerocell}[1]{-}

\definecolor{caribbeangreen}{rgb}{0.0, 0.8, 0.6}
\definecolor{brilliantlavender}{rgb}{0.96, 0.73, 1.0}
\definecolor{amethyst}{rgb}{0.6, 0.4, 0.8}
\definecolor{ao(english)}{rgb}{0.0, 0.5, 0.0}
\definecolor{arylideyellow}{rgb}{0.91, 0.84, 0.42}
\definecolor{asparagus}{rgb}{0.53, 0.66, 0.42}
\definecolor{aquamarine}{rgb}{0.5, 1.0, 0.83}
\definecolor{babyblue}{rgb}{0.54, 0.81, 0.94}
\definecolor{fwtchanged}{rgb}{0.3, 0.3, 0.7}
\definecolor{rosewood}{rgb}{0.4, 0.0, 0.04}
\definecolor{oldmauve}{rgb}{0.4, 0.19, 0.28}
\definecolor{myrtle}{rgb}{0.13, 0.26, 0.12}
\definecolor{magenta(dye)}{rgb}{0.79, 0.08, 0.48}

\definecolor{plta}{rgb}{0.12, 0.47, 0.71}
\definecolor{pltb}{rgb}{   1, 0.5, 0.05}
\definecolor{pltc}{rgb}{0.17, 0.63, 0.17}
\definecolor{pltd}{rgb}{0.84, 0.15, 0.16}

\DeclareMathOperator*{\agg}{\mathsf{agg}}
\newcommand{\Vars}{\mathcal{V}}
\newcommand{\Vals}{\mathrm{Sub}}

\newcommand{\head}{\mathrm{head}}
\newcommand{\body}{\mathrm{body}}
\newcommand{\features}[1]{\widehat{#1}}

\newcommand{\atom}{f}

\NewEnviron{smallAlign}{%
    \small
    \begin{align}
        \BODY
    \end{align}
}

\NewEnviron{smallAlign*}{%
    \small
    \begin{align*}
        \BODY
    \end{align*}
}

\NewEnviron{datalog}{%
    \small
    \begin{align*}
        \BODY
    \end{align*}
}

%% file: sections/introduction.tex
\section{Introduction}\label{sec:introduction}
Learning for planning is drawing increasing interest due to advances in deep learning architectures.
%
Most current approaches use a neural model to learn a heuristic to compute a greedy policy~\cite{staahlberg.etal.2022} or navigate a search algorithm~\cite{chen.etal.2024}.
However, such approaches cannot incorporate domain knowledge into their architecture and instead must learn all the domain mechanics from the domain definition and training data from scratch. 
It is usually the case that users already have knowledge about solving the domain, and are more interested in solution quality.
We thus propose an approach which allows users to provide the solving knowledge to the learner in order to focus on optimising solution quality, rather than relearning how to solve the planning task.

\input{figures/intro_figure_v2.tex}

\newcommand{\Condition}{\textit{Condition}}
\newcommand{\Action}{\textit{Action}}

This paper proposes incorporating domain background knowledge (BK), as common in the field of inductive logic programming~\cite{cropper.etal.2022}, into the machine learning (ML) model for planning.
%
%
We represent a generalised policy $\pi$ as an ML model which returns the best action to take in a given problem and state.
The BK we propose to provide the ML model is a generalised \emph{nondeterministic} policy $\sigma$ which returns a \emph{set} of actions to take in a given problem and state, thus restricting the hypothesis space of the model.
Such a generalised policy $\sigma$ would generally represent a satisficing but suboptimal strategy for a planning domain.
For example, a strategy for Blocksworld is to unstack all misplaced blocks onto the table in any order and then stack them back in the correct place.
We represent a generalised policy $\sigma$ as sets of lifted Datalog rules of the form $\Condition \to \Action$. 
If $\Condition$ is met in a state $s$, then the rule indicates that $\Action$ is in the policy output $\sigma(s)$.

The idea of incorporating search control knowledge with formal languages for planning is not new, as seen in works with Hierarchical Task Networks in the SHOP planner~\cite{nau.etal.1999,nau.etal.2003} and Temporal Logics in the TLPlan planner~\cite{bacchus.kabanza.2000}.
The motivation for using Datalog as search control knowledge is that (1) previous methods result in a large search space due to taking the Cartesian product of the original state space and the search control knowledge language, and (2) Datalog is an expressive language known to be P-complete for fixed programs when allowing for stratified negation~\cite{dantsin.etal.2001}.

Given a nondeterministic policy $\sigma$ encoded in Datalog, we build an ML model that learns from optimal actions of small training tasks in order to score actions from $\sigma(s)$ based on their likelihood of improving plan quality. 
We leverage \emph{Lifted Relational Neural Networks} (LRNNs) \cite{sourek.etal.2018} for this task.
LRNNs are differentiable Datalog programs which offer several advantages for planning: their inputs are relational structures such as planning states, they subsume existing relational neural architectures such as Graph Neural Networks~\cite{sourek.etal.2021}, and they naturally accept BK encoded in Datalog.
Figure~\ref{fig:intro} summarises our proposed approach.

%% file: figures/intro_figure_v2.tex
\begin{figure}[t]
    \tikzset{>=latex}
    \newcommand{\inputindent}{-0.5cm}
    \newcommand{\xgap}{2.3cm}
    \newcommand{\ygap}{0.92cm}
    \newcommand{\lineheight}{-0.075cm}
    \centering
    \def\myellipse{ellipse (0.7cm and 0.35cm)}
    \resizebox*{\columnwidth}{!}{
        \begin{tikzpicture}[scale=1]
            \tikzset{
                myrect/.style={
                    text centered,
                    minimum width=1.3cm,
                    align=center,
                    text depth=0cm,
                    rectangle,
                    font=\small
                },
                toprow/.style={
                    minimum height=0.6cm,
                    draw,
                    font=\scriptsize
                },
                model/.style={
                    minimum width=1.4cm,
                    minimum height=0.7cm,
                },
                input/.style={
                    solid,
                    draw,
                },
                taskdep/.style={
                    dashed,
                    rounded corners,
                },
                output/.style={
                    draw,
                },
                legend/.style={
                    inner sep=0cm,
                    text depth=0cm,
                    text height=0cm,
                    draw,
                    minimum width=0.525cm,
                    minimum height=0.09cm,
                    line width=0.1pt,
                }
            }



            \node[]              (oval)   at (-0.35cm+\inputindent, 1.15*\ygap) {};
            \node[legend,dashed,rounded corners=0.5mm] (dashed) at (-0.35cm+\inputindent, 0.85*\ygap) {};
            \draw[legend] (oval) ellipse (0.25cm and 0.06cm);

            \node[font=\scriptsize,anchor=west] (ovaltext)   at (-0.1cm+\inputindent, 1.15*\ygap) {generalised policy};
            \node[font=\scriptsize,anchor=west] (dashedtext) at (-0.1cm+\inputindent, 0.85*\ygap) {input/output};
            
            \node[myrect,input,taskdep] (task) at (\inputindent, -0.3*\ygap) {task};
            \node[myrect,input] (dom) at  (\inputindent,  0.3*\ygap) {domain};

            \node[myrect,model] (datalog) at (\xgap, 0) {Datalog};
            \node[myrect,model] (lrnn1) at (2*\xgap, 0) {LRNN$_{\bm{\theta}}$};
            \node[myrect,model] (lrnn2) at (3*\xgap, 0) {LRNN$_{\bm{\theta}^{*}}$\!};

            \draw (datalog) \myellipse;
            \draw (lrnn1) \myellipse;
            \draw (lrnn2) \myellipse;

            \node[myrect,toprow] (bk) at (\xgap, \ygap) {background\\[\lineheight]knowledge};
            \node[myrect,toprow] (mp) at (2*\xgap, \ygap) {message\\[\lineheight]passing};
            \node[myrect,toprow,input] (td) at (3*\xgap, \ygap) {training\\[\lineheight]data};

            \node[myrect,output,taskdep] (plan)  at (1*\xgap, -0.85*\ygap) {plan};
            \node[myrect,output,taskdep] (qplan) at (3*\xgap, -0.85*\ygap) {quality plan};

            \draw[->,taskdep] (task.east) to[out=0, in=180] ([yshift=-0.1cm]datalog.west);
            \draw[->] (dom.east)  to[out=0, in=180] ([yshift= 0.1cm]datalog.west);

            \draw[->] (bk) -- (datalog);
            \draw[->] (datalog) -- (lrnn1);
            \draw[->] (lrnn1) -- (lrnn2);
            \draw[->] (mp) -- (lrnn1);
            \draw[->] (td) -- (lrnn2);
            \draw[->,taskdep] (datalog) -- (plan);
            \draw[->,taskdep] (lrnn2) -- (qplan);
            
            
    \end{tikzpicture}
    }
    \caption{
        Outline of the proposed approach.
        %
        A domain and background knowledge is used to construct a Datalog program representing a generalised policy.
        The program can be extended with message passing rules into a parameterised LRNN trained to optimise plan quality.
        %
        %
    }
    \label{fig:intro}
\end{figure}
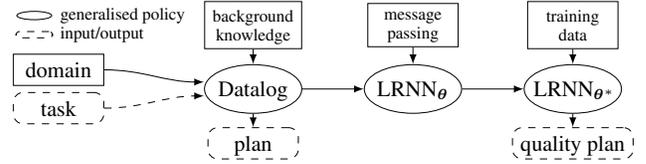

%% file: sections/background.tex
\newcommand{\problem}{\mathbf{P}}
\newcommand{\domain}{\mathbb{D}}
\newcommand{\predicates}{\mathcal{P}}
\newcommand{\objects}{\mathcal{O}}
\newcommand{\schemata}{\mathcal{A}}
\newcommand{\pred}[1]{#1}

\section{Preliminaries}\label{sec:preliminaries}
This section details the necessary preliminaries for understanding the rest of the paper.
The first two subsections introduce the formalism and representation of classical planning tasks and domains.
More specifically, planning tasks will be represented in their `lifted' form with first-order logic.
The final subsection introduces Datalog, a declarative programming language based on first-order logic.

\subsubsection{Planning Task} A planning task~\cite{geffner.bonet.2013} is a state transition model $\problem = \gen{S, A, s_0, G}$ where $S$ is a set of states, $A$ is a set of actions, $s_0 \in S$ is the initial state, and $G \subseteq S$ is a set of goal states.
An action $a \in A$ is a function $a : S \to S \cup\{\bot\}$ that maps a state $s$ to a successor $a(s) \in S$ if $a$ is applicable in $s$, otherwise $a(s) = \bot$.
%
We assume that all actions have a unit cost.
A \emph{plan} for a planning task $\problem$ is a sequence of applicable actions that transforms the initial state to a goal state when applied in order.
Formally, a plan is of the form $a_1, \ldots, a_n$ such that $s_i = a_i(s_{i-1})$ for all $i \in [n] := \{1, \ldots, n\}$ and $s_n \in G$, and we call $s_0, \ldots, s_n$ the \emph{trace} of the plan.
A plan is called \emph{optimal} if it is shortest among all plans.
A planning task is \emph{solvable} if it has at least one plan, and unsolvable otherwise.
A state $s$ is a \emph{dead-end} if the new planning task $\problem_s=\gen{S, A, s, G}$ is unsolvable.

\subsubsection{Planning Domain} In practice, planning tasks are represented in a \emph{lifted} form~\cite{lauer.etal.2021} given by a tuple $\gen{\predicates, \objects, \schemata, s_0, g}$ and set of variables $\Vars$, where $\predicates$ denotes a finite set of first-order predicates, $\objects$ a set of objects, $\schemata$ a set of action schemata, $s_0$ the initial state, and $g$ now the goal condition.
A \emph{predicate} $\pred{p} \in \predicates$ is a symbol with a corresponding arity denoting how many parameters it has.
An \emph{atomic formula} over $\Vars\cup\objects$ is an expression of the form $\pred{p}(X_1,\ldots,X_n)$ where $\pred{p}\in\predicates$ and $X_i\in\Vars\cup\objects$. 
An atomic formula where $X_i\in\objects$ for $i\in[n]$ is called a \emph{(ground) atom}. 
A \emph{substitution} is a map $v\colon\Vars\cup\objects\to\objects$ such that $v(o)=o$ for all $o\in\objects$. 
The set of all substitutions is denoted $\Vals$.
A substitution $v$ and atomic formula $\alpha=\pred{p}(X_1,\ldots,X_n)$ induces a ground atom $v(\alpha)=\pred{p}(v(X_1),\ldots,v(X_n))$.
States in a lifted planning task are represented as sets of ground atoms, and $s_0$ is the initial state.
The goal condition $g$ is also a set of ground atoms and a state $s$ is a goal state if $g \subseteq s$.

An \emph{action schema} $a \in \schemata$ in a lifted planning task is a tuple $\gen{\Vars(a), \pre(a), \add(a), \del(a)}$ where $\Vars(a)$ is a set of variables, and the preconditions $\pre(a)$, add effects $\add(a)$, and delete effects $\del(a)$ are sets of atomic formulas over $\Vars(a) \cup \objects$.
A \emph{ground action} is an action schema with all its variables substituted with objects, noting that the preconditions, add and delete effects of ground actions are sets of ground atoms.
A ground action $a$ is \emph{applicable} in a state $s$ if $\pre(a) \subseteq s$, in which case we define its \emph{successor} $a(s) = (s \setminus \del(a)) \cup \add(a)$, and $a(s) = \bot$ otherwise.
For the remainder of the paper, we assume planning tasks are represented in a lifted form.
A \emph{planning domain} is a tuple $\domain=\gen{\predicates, \schemata}$ and a planning task belongs to a domain if it shares the same predicates and schemata as $\domain$.

\newcommand{\database}{\mathbf{D}}  
\newcommand{\databaseSchema}{\mathcal{D}}
\newcommand{\databaseInstances}[1]{\mathit{Ins}({#1})}
\newcommand{\relations}[1]{\mathit{Rels}(#1)}
\newcommand{\constants}[1]{\mathit{Cons}(#1)}
\newcommand{\herbrand}[1]{\mathit{HB}(#1)}
\newcommand{\stratification}{\mathrm{str}}
\newcommand{\program}{\mathcal{F}}
\newcommand{\map}{\mathcal{M}_{\program}}
\newcommand{\mapi}[1]{\mathcal{M}_{\program_{#1}}}
\newcommand{\range}{\mathrm{rng}}

\subsubsection{Datalog}
We outline necessary definitions and results on Datalog and refer to~\cite{dantsin.etal.2001} for details. 
A \emph{literal} $\lambda$ is either an atomic formula $\a$, or its negation $\neg \a$.
The notion of a substitution $v$ is extended to literals by $v(\neg\alpha)=\neg v(\alpha)$ and to sets of literals $\Phi$ by $v(\Phi)=\{v(\alpha)\mid \alpha\in\Phi\}$.
Given a set of ground atoms $s$ and a ground literal $\lambda$, we say that $\lambda$ \emph{holds} in $s$ if $\lambda \in s$ provided $\lambda$ is a non-negated atom and $\lambda\not\in s$ otherwise.
A set of ground literals $\Phi$ holds in $s$ if all $\lambda\in \Phi$ hold in $s$, and this fact is denoted as $s\models \Phi$. 
A \emph{(Datalog) rule} $r$ is an expression
\begin{align*}
    \alpha \la \lambda_1, \ldots, \lambda_m, \quad m \geq 0
\end{align*}
where $\alpha$ is an atomic formula, and is called the \emph{head} of the rule and denoted $\head(r)$, while $\lambda_1, \ldots, \lambda_m$ is called the \emph{body} of the rule and is denoted $\body(r)$.
We say that a set of atoms $s$ is \emph{closed under} the rule $r$ if for all substitutions $v\in\Vals$, $s\models v(\body(r))$ implies $v(\head(r))\in s$.

A \emph{Datalog program} is a finite set $\program$ of Datalog rules over a given set of predicates $\predicates$ and variables $\Vars$.
A program $\program$ is \emph{stratified}~\cite{apt.blair.1991} if there exists a stratification function $\stratification: \predicates \to \N$ assigning levels to each predicate such that if a predicate $\pred{p}$ appears in the head of a rule $r$, and a predicate $\pred{q}$ appears in a literal $\lambda\in\body(r)$, then $\stratification(\pred{p}) \geq \stratification(\pred{q})$, and furthermore $\stratification(\pred{p}) \ne \stratification(\pred{q})$ if $\lambda$ is a negated atomic formula.
The stratification $\stratification$ partitions $\program=\bigcup_{i\in\N}\program_i$ where $\program_i=\{r\in\program\mid \stratification(r)=i\}$.

Given a Datalog program $\program$ and a set of ground atoms $s$ over a set of objects $\objects$, the execution of $\program$ applied to $s$ results in a set of ground atoms $\map(s)$.
The set $\map(s)$ is called the \emph{canonical model} for $\program$ and $s$. 
If $\program$ contains no negation, $\map(s)$ is the minimal set of atoms extending $s$ that is closed under the rules in $\program$. 
If $\program$ contains negations and is stratified, we define the canonical model by computing iteratively over the stratifications by $\map(s)=\mapi{n}(\mapi{n-1}(\ldots\mapi{1}(\mapi{0}(s))\ldots))$ where $n$ is the largest stratification level.
The model $\map(s)$ is unique regardless of the choice of stratification for $\program$~\cite{apt.blair.1991}.
In this paper, we focus on programs $\program$ fixed in $\map$, in which case the model $\map$ is P-complete~\cite{apt.blair.1991};~\cite[Thm. 5.1]{dantsin.etal.2001}.

%% file: sections/bk-domains.tex
\newcommand{\ag}{\texttt{ag}}
\newcommand{\ug}{\texttt{ug}}
\newcommand{\ap}{\texttt{aa}}
\newcommand{\nd}{\mathrm{nd}}

\section{Background Knowledge Policies}\label{sec:bk-domains}
In this section, we formalise how Datalog programs can be used as policies for planning domains and outline formal properties that make them suitable BK.

\subsection{Datalog Policies for Planning}

We begin by defining a general notion of a nondeterministic (ND) policy for \emph{deterministic} planning problems.
Next, we define how a Datalog program can induce such a policy for planning problems.
These Datalog programs then form the core of the BK policies utilised by the learning and planning algorithms.

\begin{definition}[Non-deterministic Policy]
    A \emph{(ND) policy} for a problem $\problem=\gen{S, A, s_0, G}$ is a function of the form $\sigma: S \to 2^A$.
    A \emph{$\sigma$-trajectory} from a state $s_1\in S$ is a finite sequence of states $s_1, \ldots, s_n$ such that for all $i=1, \ldots, n-1$, there exists an action $a \in \sigma(s_i)$ such that $s_{i+1} = a(s_i)$.
    %
\end{definition}


\newcommand{\highlight}[1]{\underline{\textbf{#1}}}
\begin{definition}[Datalog Program Induced Policy]
    Let $\problem=\gen{\predicates, \objects, \schemata, s_0, g}$ be a planning task in its lifted form, and $\program$ a Datalog program.
    %
    %
    Then $\program$ induces a ND policy for $\problem$ denoted $\sigma_{\program} : S \to 2^A$ and defined by $\sigma_{\program}(s) = \map(\sett{c(\atom, s, g) \mid \atom \in s \cup g}) \cap A$ where $A$ is the set of ground actions induced by all possible substitutions of schemata in $\schemata$ by objects in $\objects$, and $c$ maps atoms in $s\cup g$ to new atoms with predicates in $\predicates_{\ag} \cup \predicates_{\ug} \cup \predicates_{\ap}$ indicating whether an atom is an \highlight{a}chieved \highlight{g}oal, \highlight{u}nachieved \highlight{g}oal, or \highlight{a}chieved non-goal \highlight{a}tom, respectively.
    Given an atom $\atom = p(o_1, \ldots, o_n)$, we define $c(\atom, s, g)$ by
    \begin{smallAlign*}
        p_{\ag}(o_1, \ldots, o_n) & \tag*{if $\atom \in s \cap g$,} \\
        p_{\ug}(o_1, \ldots, o_n) & \tag*{if $\atom \in g \setminus s$,} \\
        p_{\ap}(o_1, \ldots, o_n) & \tag*{if $\atom \in s \setminus g$.}
    \end{smallAlign*}
\end{definition}

The function $c$ explicitly encodes goal information of the planning task into the state, and is equivalent to node colouring function of facts in the graph encoding of planning tasks proposed by~\citet{chen.etal.2024a}.

We now introduce some properties that are ideal, but not necessary, to have in BK policies in order to improve their efficiency of generating plans.
%
The first ideal property is that policies avoid dead-ends and are goal achieving, meaning that randomly executing the policy will always eventually reach a goal state.

\begin{property}[BK policies are dead-end avoiding and goal achieving]\label{ass:deadend}
    It holds that for a BK policy $\sigma$ for a solvable problem $\problem=\gen{S, A, s_0, G}$, for all $\sigma$-trajectories $s_0, \ldots, s_n$ 
    beginning from the initial state, for all $i=0, \ldots, n$, the state $s_i$ is not a dead-end.
    Furthermore, each $\sigma$-trajectory is a prefix of another $\sigma$-trajectory whose final state is a goal.
\end{property}

The second ideal property is that the policies avoid cycles and thus, on average execute faster and return better plans.

\begin{property}[BK policies are cycle-free]\label{ass:cycle}
    It holds that for a BK policy $\sigma$ for a solvable problem $\problem=\gen{S, A, s_0, G}$, for all $\sigma$-trajectories $s_0, \ldots, s_n$ 
    beginning from the initial state, for all $i=0, \ldots, n$, the state $s_i\not=s_j$ for $i\not=j$.
\end{property}

The final ideal property is that policies preserve at least one optimal plan from each state.
Note that this is a stronger property than preserving at least one optimal plan from just the initial state, but a weaker property than preserving every optimal plan.

\begin{property}[BK policies preserve optimal plans at every state]\label{ass:optimal}
    It holds that for a BK policy $\sigma$ for a solvable problem $\problem=\gen{S, A, s_0, G}$, for all solvable $s \in S$, there exists an optimal plan for $\problem_s=\gen{S, A, s, G}$ whose trace is a $\sigma$-trajectory from $s$.
\end{property}

By preserving an optimal plan at every state, we may achieve better plans on average as if a suboptimal step is made any point along the way, it is still possible to recover by choosing the optimal actions for the remaining steps.
On the other hand, by requiring to preserve at least one optimal plan, this allows us more flexibility in defining the policies while still maintaining the best possible outcome.
In practice, the reasonable satisficing strategy encoded into a BK policy will generally encapsulate most optimal plans at each state.
Although there is no systematic method to prove that an arbitrary policy for a given domain satisfies the aforementioned properties, it is possible to test for them empirically on a set of validation problems.


We also note that the properties have a connection to the notions of weak, strong, and strong-cyclic solution definitions in fully observable nondeterministic (FOND) planning~\cite{cimatti.etal.2003}.
It is possible to define a one-to-one mapping between a pair of a planning problem $\problem$ and a corresponding ND policy $\sigma$ to a ND planning problem $\problem_{\nd}$.
The main idea involves constructing exactly one nondeterministic action at each state with effects corresponding to actions in $\sigma$.
Thus, there only exists one policy $\sigma_{\nd}$ for $\problem_{\nd}$ by taking the one nondeterministic constructed action at each state.
%
%
Property~\ref{ass:deadend} (resp. \ref{ass:deadend} and \ref{ass:cycle} combined) is equivalent to strong-cyclic (resp. strong) solutions for $\problem_{\nd}$, while
Property~\ref{ass:optimal} implies weak solutions for $\problem_{\nd}$, but not the converse.

\renewcommand{\ag}[1]{#1${_\texttt{ag}}$}
\renewcommand{\ug}[1]{#1${_\texttt{ug}}$}
\renewcommand{\ap}[1]{#1${_\texttt{aa}}$}
\renewcommand{\pred}[1]{\text{#1}}
\newcommand{\derAtom}[1]{#1^{\exists}}
\newcommand{\lapre}{\la^{\!\!\!\text{pre}}}  
\newcommand{\notexists}{\neg\exists}
\renewcommand{\and}{,}
\newcommand{\qq}{\!\!\qquad}
\renewcommand{\notexists}{\not\exists}

\subsection{Example BK Policies}
As aforementioned, we conclude this section by providing examples of the BK policies. 
We introduce the following additional shorthand notations.
Let $\problem = \gen{\predicates, \objects, \schemata, s_0, g}$ be a planning problem in its lifted form.
The rule $\a \lapre \lambda_1, \ldots, \lambda_m$ where the predicate of $\a$ is an action schema $a \in \schemata$ is a shorthand\footnote{We further added object typing as defined in the PDDL domain files to the rule body but omitted this from the notation for brevity.} for $\a \la \lambda_1, \ldots, \lambda_m, \mu_1, \ldots, \mu_n$ where $\pre(a) = \sett{\mu_1, \ldots, \mu_n}$.
In the following examples, we also introduce the rules $p(X_1, \ldots, X_n) \la p_{\texttt{ag}}(X_1, \ldots, X_n)$ and $p(X_1, \ldots, X_n) \la p_{\texttt{aa}}(X_1, \ldots, X_n)$ for each $n$-ary predicate $p \in \predicates$ in the planning problem in order for the $\lapre$ rules to execute.
Furthermore, if stratified negation restrictions are followed, it is possible to derive whether any instantiation of a predicate $p$ can be derived by introducing a new nullary predicate $\derAtom{p}$ and the rule $\derAtom{p} \la p(X_1, \ldots, X_n)$.
%
We provide the PDDL domain descriptions of the examples in the appendix.

\begin{example}[Applicable Actions]
    The baseline BK policy we can provide for any planning domain is the ND policy that returns all applicable actions for each state.
    The Datalog program would be given by
    \begin{datalog}
        a(X_1,\ldots,X_n) \lapre, \quad \forall a{} \in \schemata,
    \end{datalog}
    where $\Vars(a) = \sett{X_1, \ldots, X_n}$.
    %
    We note that this policy does not satisfy Properties~\ref{ass:deadend} and \ref{ass:cycle} for all domains, but does satisfy Property~\ref{ass:optimal} by construction.
    Any other valid BK Datalog policy would be more specific as it would return a subset of applicable actions.
\end{example}

\newcommand{\wpb}{\pred{well\_placed}}
\begin{example}[Blocksworld]


    The Blocksworld domain is a well-known planning task that involves manipulating stacks of blocks to achieve a target configuration.
    Blocksworld is known to be solvable in polynomial time but NP-hard for optimal planning~\cite{gupta.nau.1992,slaney.thiebaux.2001}.
    We use the canonical polynomial BK policy of solving Blocksworld, which consists of first relocating all misplaced blocks either on the table or directly onto their goal location without disrupting other blocks, followed by relocating all remaining misplaced blocks from the table onto their goal location in order.
    \citet{slaney.thiebaux.2001} named this strategy GN1 after~\citet{gupta.nau.1992}.
    %

    We assume that problem goals fully specify the location of every block. 
    Then to encode GN1 in Datalog, we first introduce a derived predicate $\wpb(A)$, indicating that a block $A$ is well-placed if all blocks below $A$ are also well-placed and $A$ is correctly positioned in its goal location.
    \begin{datalog}
        \wpb(A) &\la \pred{\ag{on}}(A, B) \and \wpb(B) \\
        \wpb(A) &\la \pred{\ag{on\_table}}(A)
    \end{datalog}

    Then the relevant action rules are as follows.
    \begin{datalog} 
            \pred{unstack}(A, B) &\lapre \neg\wpb(A) \tag{B1} \\
            \pred{stack}(A, B) &\lapre \pred{\ug{on}}(A, B) \and \wpb(B) \tag{B2} \\
            \pred{pickup}(A) &\lapre \pred{\ug{on}}(A, B) \and \wpb(B) \and \pred{clear}(B) \tag{B3} \\
            \pred{putdown}(A) &\lapre \pred{\ug{on}}(A, B) \and \neg \wpb(B) \tag{B4} \\
            \pred{putdown}(A) &\lapre \pred{\ug{on\_table}}(A)\tag{B5} \\
            \pred{putdown}(A) &\lapre \pred{\ug{on}}(A, B) \and \pred{\ap{on}}(C, B) \tag{B6}
    \end{datalog}

    Rule (B1) unstacks any block $A$ that is not well-placed. Subsequently, there are two possible actions: (i) stack $A$ on $B$ if $A$'s goal position is on $B$ and $B$ is well-placed (B2), or (ii) put it on the table if $B$ is not well-placed (B4), $A$'s goal position is on the table (B5), or there is another block on top of $B$ (B6).
    Once there are no unstack actions left, all that remains is to pick up from the table any block $A$ guaranteed to have a goal position on another block $B$ that is well-placed (B3).
    
    Given that Blocksworld has no dead-ends, this policy satisfies Property~\ref{ass:deadend}.
    Furthermore, Property~\ref{ass:cycle} is guaranteed as there are no redundant actions.
    Property~\ref{ass:optimal} is also satisfied, as one method for computing optimal Blocksworld plans involves correctly choosing which misplaced block to put on the table by computing the minimal hitting set of deadlocks in a state~\cite{slaney.thiebaux.2001} in the GN1 algorithm.
\end{example}

\newcommand{\insconfig}{\pred{ins\_config}}
\begin{example}[Satellite]\label{eg:satellite}
    The Satellite domain consists of a set of satellites, each of which contains some set of imaging instruments.
    Each instrument supports a specific imaging mode, and must be calibrated by pointing it in a specific direction.
    Each satellite can only power on one instrument at a time.
    A problem from the domain involves taking images under certain modes of different directions in the sky, followed by pointing the satellites in specific directions.

    Satellite is NP-hard to optimise but solvable in polynomial time with a natural greedy strategy that is 6-approximating~\cite{helmert.etal.2006}.
    We implement this greedy strategy as the BK Datalog policy.
    It involves repeatedly performing the subroutine of (1) switching on an instrument that may contribute to a goal image, (2) pointing the corresponding satellite in the direction of the calibration target and (3) calibrating it, (4) turning towards a goal direction and (5) taking an image.
    Then we turn all satellites to their corresponding goal positions.
    One may further switch off any instrument after use if another instrument in the same satellite is needed, but for the tested problems this is not required due to the abundance of available satellites.
    We first introduce the derived predicate $\insconfig(S, I, M, D)$ as a macro for a conjunction of atomic formulae specifying that a satellite $S$ contains an instrument $I$ supporting mode $M$, and is a candidate for taking a goal image with mode $M$ at direction $D$.
    \begin{datalog}
        \insconfig(S, I, M, D_g) &\la 
        \pred{supports}(I, M) \and 
        \pred{on\_board}(I, S) \and \\ &\qq
        \pred{\ug{have\_image}}(D_g, M)
    \end{datalog}

    Then the relevant action rules are as follows.
    \begin{datalog}
        \pred{switch\_on}(I, S) &\lapre 
        \insconfig(S, I, M, D_g) \tag{S1} \\
        \pred{turn\_to}(S, D_n, D_p) &\lapre
        \insconfig(S, I, M, D_g) \and \\ &\qq 
        \pred{calibration\_target}(I, D_n) \and \\ &\qq 
        \pred{power\_on}(I) \and 
        \neg\pred{calibrated}(I) \and \\ &\qq 
        \neg\derAtom{\pred{calibrate}} \and 
        \neg\derAtom{\pred{take\_image}} \tag{S2} \\
        \pred{calibrate}(S, I, D) &\lapre
        \insconfig(S, I, M, D_g) \and \\ &\qq 
        \neg \pred{calibrated}(I) \tag{S3} \\
        \pred{turn\_to}(S, D_n, D_p) &\lapre
        \insconfig(S, I, M, D_g) \and \\ &\qq 
        \pred{calibrated}(I) \and \\ &\qq
        \neg\derAtom{\pred{calibrate}} \and 
        \neg\derAtom{\pred{take\_image}} \tag{S4} \\
        \pred{take\_image}(S, I, M, D) &\lapre
        \insconfig(S, I, M, D) \tag{S5} \\
        \pred{turn\_to}(S, D_n, D_p) &\lapre
        \neg\text{have\_image}_{\texttt{ug}}^{\exists} \and 
        \pred{\ug{pointing}}(S, D_n) \tag{S6}
    \end{datalog}

    Rule (S1) involves switching on an instrument that may help to take a goal image with the aid of the derived $\insconfig$ predicate.
    Rules (S2) and (S4) determine whether to turn a satellite to a calibration or goal image direction, respectively, with body atoms $\neg\derAtom{\pred{calibrate}}$ and $\neg\derAtom{\pred{take\_image}}$ ensuring that the turn\_to actions are prioritised last.
    This is done in order to avoid loops as per Property~\ref{ass:cycle} by ensuring that each turn\_to action has a meaning, whether that is to allow for a satellite to calibrate its instrument, or to take a goal image.
    Rules (S3) and (S5) determine whether to calibrate or take an image, respectively, again with the $\insconfig$ predicate ensuring that each calibration or take image action contributes towards a goal.
    Lastly, (S6) determines to turn satellites to their goal directions, with the body atom $\neg\text{have\_image}_{\texttt{ug}}^{\exists}$ ensuring that these actions are only done once all images have been taken.

    We note that this BK policy satisfies both Properties 1 and 2.
    There are no dead-ends in a solvable problem in the described Satellite domain.
    Furthermore, loops do not occur as each action progresses a subroutine towards achieving a goal.
    However, the policy does not satisfy Property 3.
    Specifically, it is sometimes optimal to turn a satellite even if a calibrate action is derivable.
    This suboptimality arises from attempting to encode a cycle-free policy with the negative atoms in (S2) and (S4).
    The fact that the policy does not preserve optimal solutions can be discovered by executing it on states with explicitly precomputed optimal actions.
    In practice, we found that the set of derived actions does not contain any optimal action in 0.11\% of the tested states.
\end{example}

%% file: sections/lrnn.tex
\newcommand{\smset}[1]{\{\mskip-5mu\{#1\}\mskip-5mu\}}
\section{LRNN Datalog Program}
\emph{Lifted Relational Neural Networks} (LRNN)~\cite{sourek.etal.2018} do not have fixed computation structures, as in usual deep learning architectures, and define them declaratively via logic programming.
They bring more expressiveness for learning with structured data, while subsuming existing neural architectures like convolutional, recurrent, or graph neural networks~\cite{sourek.etal.2021}.
In this section, we introduce the general LRNN principle, and then discuss how to instantiate a LRNN given a BK policy and a planning domain.
Figure~\ref{fig:lrnn} summarises the LRNN concept.



\input{figures/lrnn_figure.tex}

\subsubsection{General LRNN Architecture}
LRNNs can essentially be viewed as differentiable Datalog programs $\program$, endowing the contained rules with tuples of learnable parameters. 
%
%
An input to an LRNN is a set of ground atoms $s$ with each $\atom\in s$ associated (optionally) with a feature vector $\features{\atom}$.
The output is then the canonical model $\map(s)$ of $\program$ with an embedding vector associated with each derived ground atom in $\map(s)$.

Specifically, an LRNN assigns every rule $r \in \program$ a matrix $\bm{H}^r$ associated with its $\head(r)$, and matrices $\bm{B}^r_\lambda$ associated with each of its body atoms $\lambda \in \body(r)$.
%
Given an input $s$, it then recursively computes vectors $\features{\atom}$ for each ground atom $\atom \in \map(s) \setminus s$ based on the derivation of $\atom$ by $\program$.
The computation of $\features{\atom}$ is done in two steps.

Firstly, we consider all instances of a single rule $r$ that can derive $\atom$.  
For each $r\in\program$ and derivable $\atom$, we define a restricted set of substitutions
\begin{smallAlign*}
    \Vals_{r,\atom}=\{v\in \Vals\mid v(\head(r))=\atom \wedge \map(s)\models v(\body(r))\}
\end{smallAlign*}
and compute a multiset of ``messages'' by
\begin{smallAlign*}
    M_{r,\atom}=
    \msetbig{
        \phi_1
        \Big(\textstyle\sum_{\lambda\in\body(r)}\bm{B}^r_{\lambda}\cdot\features{v(\lambda)}\Big)
        \mid 
        v\in \Vals_{r,\atom}
    }
\end{smallAlign*}
where $\phi_1$ is an activation function like the sigmoid, tanh, or ReLU applied component-wise.
Secondly, we define the set of rules that can generate $\atom$ by
$
    \program_\atom=\{r\in\program\mid \Vals_{r,\atom}\ne\emptyset\}
$
and combine the message multisets $M_{r,\atom}$ for all $r\in\program_\atom$ by
\begin{smallAlign*}
    \features{\atom}=\phi_2\Big(\textstyle\sum_{r\in\program_\atom}\bm{H}^r\cdot \agg(M_{r,\atom})\Big)
\end{smallAlign*}
where $\phi_2$ is another activation function, and $\agg$ is an aggregation function such as a component-wise summation, mean, or maximum.

\subsubsection{LRNNs for Planning}
Given a BK Datalog policy $\program$ for a planning domain $\domain$, we extend it with a simple ``message passing'' scheme~\cite{gilmer.etal.2017}, instantiating an extended Datalog program $\program'$.
%
%
Firstly, for each $n$-ary predicate $p\in\predicates$ in the planning problem, we introduce an extra rule
\begin{datalog}
    \pred{n-ary}(X_1,\ldots,X_n) &\la p(X_1,\ldots,X_n)
\end{datalog}
mapping all atoms of the same arity to atoms with the same predicate for simplicity.
This then allows to define very generic rules representing message passing between the domain objects. 
To follow its standard binary form, we further introduce the notion of ``edges'' by adding a rule for each arity $n\geq 2$ and $i,j\in[n]$, $i\neq j$ as follows.
\begin{datalog}
    \pred{edge}(X_i, X_j) &\la \pred{n-ary}(X_1,\ldots,X_n).
\end{datalog}
Next, we introduce object embeddings with predicates $h_0, \ldots, h_{L}$, where $L$ denotes the number of message passing ``layers''.
In the initial layer, we aggregate the object embeddings directly from all the corresponding $n$-ary atoms' positions $i\in[n]$ with
\begin{datalog}
    {h_0}(X_i) &\la \pred{n-ary}(X_1,\ldots,X_n),
\end{datalog}
and in the subsequent layers, we update the embeddings through the defined edges in the standard (GNN) fashion as
\begin{datalog}
    {h_{i+1}}(Y) &\la {h_i}(X), \pred{edge}(X,Y) \\
    {h_{i+1}}(Y) &\la {h_i}(Y)
\end{datalog}
%
Finally, we extend the body of each rule $r\in\program$ with these embeddings into
$\body(r)\cup\{{h}_k(X)\mid X\text{ occurs in }r\}.$

%% file: figures/lrnn_figure.tex
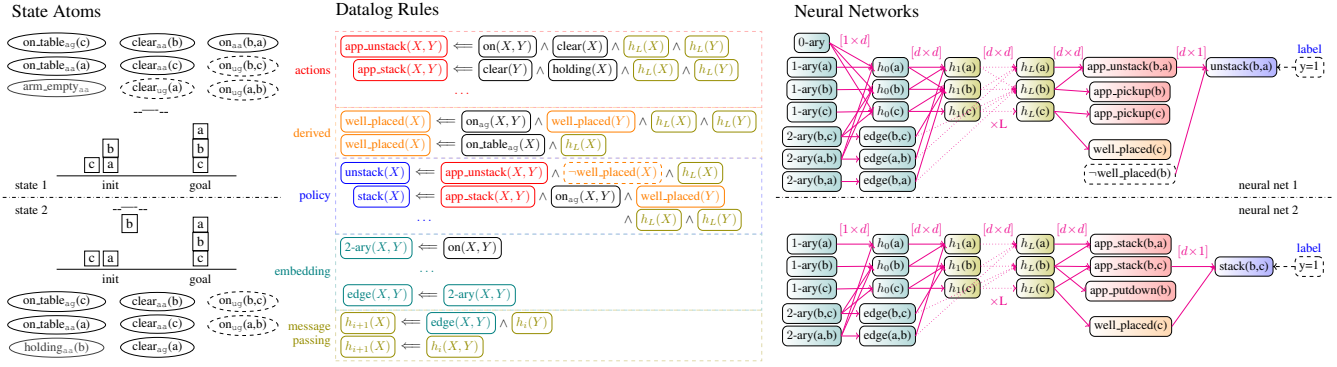
\begin{figure*}[ht!]
	\centering
	\resizebox*{\textwidth}{!}{
        \begin{tikzpicture}
            [
            ar/.style={->,>=latex},
            mynode/.style={
                draw, scale = 1.0,  minimum size=1cm, rounded corners,left color=white,
                minimum height=1cm,
                align=center
            }
            ]

            \newcommand{\textycoord}{14.5cm}
            \newcommand{\splitlinecoord}{9.3cm}

            \node[] at (-5cm,\textycoord) {\LARGE{State Atoms}};
            \node[] at (4.2cm,\textycoord) {\LARGE{Datalog Rules}};
            \node[] at (17.25cm,\textycoord) {\LARGE{Neural Networks}};

            \draw[dash dot] (-6.5cm,\splitlinecoord) -- 
            node[below left=0.1cm and 2.5cm] {state 2} 
            node[above left=0.1cm and 2.5cm] {state 1} (1.25cm,\splitlinecoord);

            \draw[dash dot] (15cm,\splitlinecoord) -- 
            node[align=left, below right=0.1cm and 5cm] {neural net 2} 
            node[align=left, above right=0.1cm and 5cm] {neural net 1} (30.5cm,\splitlinecoord);
            
            
            \begin{scope}[xshift={-3.5cm}]
                \newcommand{\blocksFontSize}{\large}
                \newcommand{\size}{0.4cm}
                \newcommand{\linegap}{-0.3cm}
	            
	            \begin{scope}[yshift=5.2cm]
	            	\newcommand{\xgap}{2.5cm}
	            	\newcommand{\ygap}{5cm}
	            	\tikzstyle{box} = [draw, rectangle, minimum size=\size]
	            	
	            	\node[box] (a1) at (0,\ygap) {\blocksFontSize a};
					\node[box, above=0cm of a1, label={[label distance=0.7cm]60:{\_\_----\_\_}}] (b1) {\blocksFontSize b};
					\node[box, left=0.1cm of a1] (c1) {\blocksFontSize c};
					
					\node[box] (c1') at (\xgap,\ygap) {\blocksFontSize c};
					\node[box, above=0cm of c1'] (b1') {\blocksFontSize b};
					\node[box, above=0cm of b1'] (a1') {\blocksFontSize a};
					
					\draw (-1.5cm,\ygap \linegap) -- (\xgap+1cm,\ygap \linegap); 
					\node[below=0.1cm of a1] {init};      
					\node[below=0.1cm of c1'] {goal}; 
	            	
	           	\end{scope}
            	
	            \begin{scope}[yshift=7.6cm]
	            	\newcommand{\xgap}{2.5cm}
	            	\newcommand{\ygap}{5cm}
	            	\tikzstyle{box} = [draw, rectangle, minimum size=\size]
	            	\node[box] (a2) at (0,0) {\blocksFontSize a};
	            	\node[box, above right=0.5cm and 0.1cm of a2, label={\_\_----\_\_}] (b2) {\blocksFontSize b};
	            	\node[box, left=0.1cm of a2] (c2) {\blocksFontSize c};
	            	
	            	\node[box] (c2') at (\xgap,0) {\blocksFontSize c};
	            	\node[box, above=0cm of c2'] (b2') {\blocksFontSize b};
	            	\node[box, above=0cm of b2'] (a2') {\blocksFontSize a};
	            	
	            	\draw (-1.5cm,\linegap) -- (\xgap+1cm,\linegap);
	            	\node[below=0.1cm of a2] {init};      
	            	\node[below=0.1cm of c2'] {goal}; 
	            \end{scope}
            \end{scope}
           	
            \tikzstyle{fact} = [
                draw, 
                ellipse, 
                inner sep=0.03cm,
            ]
           	
            \begin{scope}[xshift={-5cm}, yshift=13.6cm]
                \newcommand{\xgap}{2.8cm}
                \newcommand{\ygap}{0.1cm}
                
                \node[fact] (f1) at (0,0) {\pred{\ag{on\_table}}(c)};
                \node[fact, below=\ygap of f1] (f2) {\pred{\ap{on\_table}}(a)};
                \node[fact, darkgray, below=\ygap of f2] (f3) {\pred{\ap{arm\_empty}}};
                
                \node[fact] (f4) at (\xgap,0) {\pred{\ap{clear}}(b)};
                \node[fact, below=\ygap of f4] (f5) {\pred{\ap{clear}}(c)};
                \node[fact, dashed, below=\ygap of f5] (f6) {\pred{\ug{clear}}(a)};
                
                \node[fact] (f7) at (1.85*\xgap,0) {\pred{\ap{on}}(b,a)};
                \node[fact, dashed, below=\ygap of f7] (f8) {\pred{\ug{on}}(b,c)};
                \node[fact, dashed, below=\ygap of f8] (f9) {\pred{\ug{on}}(a,b)};
                
            \end{scope}
           	
			\begin{scope}[xshift={-5cm}, yshift=6.4cm]
				\newcommand{\xgap}{2.8cm}
				\newcommand{\ygap}{0.1cm}
				
				\node[fact] (f1') at (0,0) {\pred{\ag{on\_table}}(c)};
				\node[fact, below=\ygap of f1'] (f2') {\pred{\ap{on\_table}}(a)};
				\node[fact, darkgray, below=\ygap of f2'] (f3') {\pred{\ap{holding}}(b)};
				
				\node[fact] (f4') at (\xgap,0) {\pred{\ap{clear}}(b)};
				\node[fact, below=\ygap of f4'] (f5') {\pred{\ap{clear}}(c)};
				\node[fact, below=\ygap of f5'] (f6') {\pred{\ag{clear}}(a)};
				
				\node[fact, dashed] (f7') at (1.85*\xgap,0) {\pred{\ug{on}}(b,c)};
				\node[fact, dashed, below=\ygap of f7'] (f8') {\pred{\ug{on}}(a,b)};
				
			\end{scope}

            \newcommand{\datalogxleft}{-0.2cm}
            \newcommand{\xgap}{-0.1cm}
            \newcommand{\ygap}{-0.415cm}
            \newcommand{\boxygap}{-0.775cm}
            \newcommand{\hacka}{0.0cm}
            \newcommand{\hackb}{0.0cm}

            \begin{scope}[xshift={6.5cm+\datalogxleft}, yshift=13.5cm]
				\tikzstyle{atom} = [draw, rectangle, rounded corners, minimum size=.5cm]
				
				
				\node[circle] (impla1) at (0,0) {$\impliedby$};
				\node[atom, red, left=\xgap of impla1] (ra1h) {\pred{app\_unstack}$(X,Y)$};
				\node[atom, right=\xgap of impla1] (ra1b1) {\pred{on}$(X,Y)$};
				\node[circle, right=\xgap of ra1b1] (conja1) {$\wedge$};
				\node[atom, right=\xgap of conja1] (ra1b1) {\pred{clear}$(X)$};
				\node[circle, right=\xgap of ra1b1] (conja12) {$\wedge$};
				\node[atom, olive, right=\xgap of conja12] (ra1h1) {{$h_{L}(X)$}};
				\node[circle, right=\xgap of ra1h1] (conja13) {$\wedge$};
				\node[atom, olive, right=\xgap of conja13] (ra1h2) {{$h_{L}(Y)$}};
				
				\node[circle, below=\ygap of impla1] (impla2) {$\impliedby$};
				\node[atom, red, left=\xgap of impla2] (ra2h) {\pred{app\_stack}$(X,Y)$};
				\node[atom, right=\xgap of impla2] (ra2b1) {\pred{clear}$(Y)$};
				\node[circle, right=\xgap of ra2b1] (conja2) {$\wedge$};
				\node[atom, right=\xgap of conja2] (ra2b2) {\pred{holding}$(X)$};
				\node[circle, right=\xgap of ra2b2] (conja12) {$\wedge$};
				\node[atom, olive, right=\xgap of conja12] (ra2h1) {{$h_{L}(X)$}};
				\node[circle, right=\xgap of ra2h1] (conja13) {$\wedge$};
				\node[atom, olive, right=\xgap of conja13] (ra2h2) {{$h_{L}(Y)$}};
				
                \node[circle, below=\ygap of impla2] (dots) {\phantom{$\impliedby$}};
				\node[circle] at (dots) {\textcolor{red}{$\dots$}};
				
				\node[circle] (impl1) at ([xshift=-0.475cm, yshift=\boxygap]dots) {$\impliedby$};  
				\node[atom, orange, left=\xgap of impl1] (r1h) {\wpb$(X)$};
				\node[atom, right=\xgap of impl1] (r1b1) {\pred{\ag{on}$(X,Y)$}};
				\node[circle, right=\xgap of r1b1] (conj) {$\wedge$};
				\node[atom, orange, right=\xgap of conj] (r1b2){\wpb$(Y)$};
				\node[circle, right=\xgap of r1b2] (conja12) {$\wedge$};
				\node[atom, olive, right=\xgap of conja12] (ra1h1) {{$h_{L}(X)$}};
				\node[circle, right=\xgap of ra1h1] (conja13) {$\wedge$};
				\node[atom, olive, right=\xgap of conja13] (ra1h22) {{$h_{L}(Y)$}};

				\node[circle, below=\ygap of impl1] (impl2) {$\impliedby$};
				\node[atom, orange, left=\xgap of impl2] (r2h)  {\pred{well\_placed}$(X)$};
				\node[atom, right=\xgap of impl2] (r2b1) {\pred{\ag{on\_table}$(X)$}};
				\node[circle, right=\xgap of r2b1] (conja12) {$\wedge$};
				\node[atom, olive, right=\xgap of conja12] (ra1h1) {{$h_{L}(X)$}};

				
				
				\node[circle] (impl3) at ([xshift=-0.575cm, yshift=\boxygap+\hacka]impl2)  {$\impliedby$};  
				\node[atom, blue, left=\xgap of impl3] (r3h) {\pred{unstack}$(X)$};
				\node[atom, red, right=\xgap of impl3] (r3b1) {\pred{app\_unstack}$(X,Y)$};
				\node[circle, right=\xgap of r3b1] (conja11) {$\wedge$};
				\node[atom, dashed, orange, right=\xgap of conja11] (r3b2) {$\neg$\wpb$(X)$};
				\node[circle, right=\xgap of r3b2] (conja12) {$\wedge$};
				\node[atom, olive, right=\xgap of conja12] (ra1h1) {{$h_{L}(X)$}};
				
				\node[circle, below=\ygap of impl3] (impl4) {$\impliedby$};
				\node[atom, blue, left=\xgap of impl4] (r4h) {\pred{stack}$(X)$};
				\node[atom, red, right=\xgap of impl4] (r4a1) {\pred{app\_stack}$(X,Y)$};
				\node[circle, right=\xgap of r4a1] (conja21) {$\wedge$};
				\node[atom, right=\xgap of conja21] (r4b1) {\pred{\ag{on}$(X,Y)$}};
				\node[circle, right=\xgap of r4b1] (conj4) {$\wedge$};
				\node[atom, orange, right=\xgap of conj4] (r4b2){\wpb$(Y)$};

				\node[circle, below=0.0cm of conj4] (conja12) {$\wedge$};
				\node[atom, olive, right=\xgap of conja12] (ra1h1) {{$h_{L}(X)$}};
				\node[circle, right=\xgap of ra1h1] (conja13) {$\wedge$};
				\node[atom, olive, right=\xgap of conja13] (ra4h2) {{$h_{L}(Y)$}};
				
				\node[circle, below=\ygap of impl4] (dots1) {\phantom{$\impliedby$}};
				\node[circle] at (dots1) {\textcolor{blue}{$\dots$}};
				
				
				\node[circle] (impl5) at ([xshift=0.075cm, yshift=\boxygap]dots1) {$\impliedby$};  
				\node[atom, teal, left=\xgap of impl5] (r5h) {\pred{2-ary}$(X,Y)$};
				\node[atom, right=\xgap of impl5] (r5b1) {\pred{{on}$(X,Y)$}};
				
				\node[circle, below=\ygap of impl5] (dots2) {\phantom{$\impliedby$}};
				\node[circle] at (dots2) {\textcolor{teal}{$\dots$}};
				
				\node[circle, below=\ygap of dots2] (impl6) {$\impliedby$};
				\node[atom, teal, left=\xgap of impl6] (r6h) {\pred{edge}$(X,Y)$};
				\node[atom, teal, right=\xgap of impl6] (r6b1) {\pred{2-ary}$(X,Y)$};
				
				
				\node[circle] (impl7) at ([xshift=-0.475cm, yshift=\boxygap+\hackb]impl6) {$\impliedby$};  
				\node[atom, olive, left=\xgap of impl7] (r7h) {{$h_{i+1}$}$(X)$};
				\node[atom, teal, right=\xgap of impl7] (r7b1) {\pred{edge}$(X,Y)$};
				\node[circle, right=\xgap of r7b1] (conj7) {$\wedge$};
				\node[atom, olive, right=\xgap of conj7] (r7b2){{$h_{i}$}$(Y)$};
				
				\node[circle, below=\ygap of impl7] (impl8) {$\impliedby$};
				\node[atom, olive, left=\xgap of impl8] (r8h) {{$h_{i+1}$}$(X)$};
				\node[atom, olive, right=\xgap of impl8] (r8b1) {{$h_{i}$}$(X,Y)$};
				
				
			\end{scope}
			

            \newcommand{\groupx}{2.95cm}
            \newcommand{\groupwidth}{11.85cm}
            \newcommand{\ruley}{0.705cm}
            \newcommand{\topgroupy}{13.925cm}
            \newcommand{\bottrim}{0.0}

			\node [draw=red!30, dashed, inner sep=0.1cm, label={[red]left:actions},
                minimum width=\groupwidth,
                minimum height=3*\ruley,
                anchor=north west
            ] (actions) 
            at (\groupx+\datalogxleft, \topgroupy) {};

			\node [draw=orange!30, dashed, inner sep=0.1cm, label={[orange]left:derived},
            minimum width=\groupwidth,
            minimum height=2*\ruley
            ] (derived) 
            at ([yshift=-2.5*\ruley]actions) {};

			\node [draw=blue!30, dashed , inner sep=0.1cm, label={[blue]left:policy},
            minimum width=\groupwidth,
            minimum height=3*\ruley
            ] (policy) 
            at ([yshift=-2.5*\ruley]derived) {};

			\node [draw=teal!30, dashed , inner sep=0.1cm, label={[teal]left:embedding},
            minimum width=\groupwidth,
            minimum height=3*\ruley,
            ] (embedding) 
            at ([yshift=-3*\ruley]policy) {};

			\node [draw=olive!30, dashed , inner sep=0.1cm, label={[olive, align=right]left:message\\passing},
            minimum width=\groupwidth,
            minimum height=(2-\bottrim)*\ruley,
            ] (gnn) 
            at ([yshift=-(2.5-0.5*\bottrim)*\ruley]embedding) {};

            \begin{scope}[xshift={16cm}]
                \renewcommand{\xgap}{1cm}
                \renewcommand{\ygap}{0.06cm}
                \tikzstyle{atom} = [draw, rectangle, rounded corners, minimum size=.5cm]
                
                \tikzstyle{neuron}  =  [rectangle, rounded corners, draw, scale = 1.0]
                \tikzstyle{ruleneuron}  =  [neuron, left color=white]
                \tikzstyle{empty}  =  [neuron, dashed]
                
                \tikzstyle{edge} = [->, magenta, out=west,in=east]  
                
                \begin{scope}[yshift=13.6cm, xshift=0cm]
                        
                        \node[ruleneuron, right color=teal!30!white] (0-ary) at (0,0) {\pred{$0$-ary}};
                        
                        \node[ruleneuron, right color=teal!30!white, below=2*\ygap of 0-ary] (1-ary-a) {\pred{$1$-ary}(a)};
                        \node[ruleneuron, right color=teal!30!white, below=\ygap of 1-ary-a] (1-ary-b) {\pred{$1$-ary}(b)};
                        \node[ruleneuron, right color=teal!30!white, below=\ygap of 1-ary-b] (1-ary-c) {\pred{$1$-ary}(c)};
                        
                        \node[ruleneuron, right color=teal!30!white, below=2*\ygap of 1-ary-c] (2-ary-bc) {\pred{$2$-ary}(b,c)};
                        \node[ruleneuron, right color=teal!30!white, below=\ygap of 2-ary-bc] (2-ary-ab) {\pred{$2$-ary}(a,b)};
                        \node[ruleneuron, right color=teal!30!white, below=\ygap of 2-ary-ab] (2-ary-ba) {\pred{$2$-ary}(b,a)};
                        
                        
                        \node[ruleneuron, right color=teal!30!white, right=\xgap of 1-ary-a] (h0-a) {{$h_0$}(a)};
                        \node[ruleneuron, right color=teal!30!white, below=\ygap of h0-a] (h0-b) {{$h_0$}(b)};
                        \node[ruleneuron, right color=teal!30!white, below=\ygap of h0-b] (h0-c) {{$h_0$}(c)};
                        
                        \node[ruleneuron, right color=teal!30!white, right=0.5*\xgap of 2-ary-bc] (edge-bc) {\pred{edge}(b,c)};
                        \node[ruleneuron, right color=teal!30!white, right=0.5*\xgap of 2-ary-ab] (edge-ab) {\pred{edge}(a,b)};
                        \node[ruleneuron, right color=teal!30!white, right=0.5*\xgap of 2-ary-ba] (edge-ba) {\pred{edge}(b,a)};
                        
                            
                            \draw[edge] (0-ary.east) -> node[above right=0.1cm and -0.45cm] {\textcolor{magenta}{$[1\!\times\!d]$}} (h0-a.west);
                            \draw[edge] (0-ary.east) -> (h0-b.west);
                            \draw[edge] (0-ary.east) -> (h0-c.west);
                            
                            \draw[edge] (1-ary-a.east) -> (h0-a.west);
                            \draw[edge] (1-ary-b.east) -> (h0-b.west);
                            \draw[edge] (1-ary-c.east) -> (h0-c.west);
                            
                            \draw[edge] (2-ary-bc.east) -> (h0-b.west);
                            \draw[edge] (2-ary-bc.east) -> (h0-c.west);
                            \draw[edge] (2-ary-ab.east) -> (h0-a.west);
                            \draw[edge] (2-ary-ab.east) -> (h0-b.west);
                            
                            \draw[edge] (2-ary-bc.east) -> (edge-bc.west);
                            \draw[edge] (2-ary-ab.east) -> (edge-ab.west);
                            \draw[edge] (2-ary-ba.east) -> (edge-ba.west);
                        
                        
                        \node[ruleneuron, right color=olive!30!white, right=\xgap of h0-a] (h1-a) {{$h_1$}(a)};
                        \node[ruleneuron, right color=olive!30!white, below=\ygap of h1-a] (h1-b) {{$h_1$}(b)};
                        \node[ruleneuron, right color=olive!30!white, below=\ygap of h1-b] (h1-c) {{$h_1$}(c)};
                        
                            
                            \draw[edge] (h0-a.east) -> node[above=0.1cm] {\textcolor{magenta}{$[d\!\times\!d]$}} (h1-a.west);
                            \draw[edge] (h0-b.east) -> (h1-b.west);
                            \draw[edge] (h0-c.east) -> (h1-c.west);
                            
                            \draw[edge] (h0-c.east) -> (h1-b.west);
                            \draw[edge] (edge-bc.east) -> (h1-b.west);
                            \draw[edge] (h0-b.east) -> (h1-a.west);
                            \draw[edge] (edge-ab.east) -> (h1-a.west);
                            \draw[edge] (h0-a.east) -> (h1-b.west);
                            \draw[edge] (edge-ba.east) -> (h1-b.west);
                        
                        
                        \node[ruleneuron, right color=olive!30!white, right=\xgap of h1-a] (hi-a) {{$h_{L}$}(a)};
                        \node[ruleneuron, right color=olive!30!white, below=\ygap of hi-a] (hi-b) {{$h_{L}$}(b)};
                        \node[ruleneuron, right color=olive!30!white, below=\ygap of hi-b] (hi-c) {{$h_{L}$}(c)};
                        
                            
                            \draw[edge, dotted] (h1-a.east) -> node[above=0.1cm] {\textcolor{magenta}{$[d\!\times\!d]$}}  (hi-a.west);
                            \draw[edge, dotted] (h1-b.east) -> (hi-b.west);
                            \draw[edge, dotted] (h1-c.east) -> node[below=0.1cm] {$\times$L} (hi-c.west);
                            
                            \draw[edge, dotted] (h1-c.east) -> (hi-b.west);
                            \draw[edge, dotted] (edge-bc.east) -> (hi-b.west);
                            \draw[edge, dotted] (h1-b.east) -> (hi-a.west);
                            \draw[edge, dotted] (edge-ab.east) -> (hi-a.west);
                            \draw[edge, dotted] (h1-a.east) -> (hi-b.west);
                            \draw[edge, dotted] (edge-ba.east) -> (hi-b.west);
                        
                        
                        \node[ruleneuron, right color=red!30!white, right=0.8*\xgap of hi-a] (app-unstack-ba) {\pred{app\_unstack}(b,a)};
                        \node[ruleneuron, right color=red!30!white, below=2*\ygap of app-unstack-ba] (app-pickup-b) {\pred{app\_pickup}(b)};
                        \node[ruleneuron, right color=red!30!white, below=\ygap of app-pickup-b] (app-pickup-c) {\pred{app\_pickup}(c)};
                        
                            \draw[edge] (hi-a.east) -> node[above=0.1cm] {\textcolor{magenta}{$[d\!\times\!d]$}}  (app-unstack-ba.west);
                            \draw[edge] (hi-b.east) -> (app-unstack-ba.west);
                            \draw[edge] (hi-b.east) -> (app-pickup-b.west);
                            \draw[edge] (hi-c.east) -> (app-pickup-c.west);

                        \node[ruleneuron, right color=orange!30!white, below=7*\ygap of app-pickup-c] (wpb-c) {\wpb(c)};
                        \node[empty, below=2*\ygap of wpb-c] (not-wpb-b) {$\neg$\wpb(b)};
                        
                        \draw[edge] (hi-c.east) -> (wpb-c.west);
                        
                        
                        \node[ruleneuron, right color=blue!30!white, right=0.8\xgap of app-unstack-ba] (unstack-ba) {\pred{unstack}(b,a)};
                        
                        \draw[edge] (app-unstack-ba.east) -> node[above=0.1cm] {\textcolor{magenta}{$[d\!\times\!1]$}}  (unstack-ba.west);
                        \draw[edge] (not-wpb-b.east) -> (unstack-ba.west);
                        
                        
                        \node[empty, right=0.5\xgap of unstack-ba, label={[blue]above:label}] (y1) {y=1};
                        \draw[dashed, ->] (y1.west) -> (unstack-ba.east);
                    
                \end{scope}  
                
                \begin{scope}[yshift=8cm]
                    
                    
                    \node[ruleneuron, right color=teal!30!white] at (0,0) (1-ary-a) {\pred{$1$-ary}(a)};
                    \node[ruleneuron, right color=teal!30!white, below=\ygap of 1-ary-a] (1-ary-b) {\pred{$1$-ary}(b)};
                    \node[ruleneuron, right color=teal!30!white, below=\ygap of 1-ary-b] (1-ary-c) {\pred{$1$-ary}(c)};
                    
                    \node[ruleneuron, right color=teal!30!white, below=2*\ygap of 1-ary-c] (2-ary-bc) {\pred{$2$-ary}(b,c)};
                    \node[ruleneuron, right color=teal!30!white, below=\ygap of 2-ary-bc] (2-ary-ab) {\pred{$2$-ary}(a,b)};
                    
                    
                    \node[ruleneuron, right color=teal!30!white, right=\xgap of 1-ary-a] (h0-a) {{$h_0$}(a)};
                    \node[ruleneuron, right color=teal!30!white, below=\ygap of h0-a] (h0-b) {{$h_0$}(b)};
                    \node[ruleneuron, right color=teal!30!white, below=\ygap of h0-b] (h0-c) {{$h_0$}(c)};
                    
                    \node[ruleneuron, right color=teal!30!white, right=0.5*\xgap of 2-ary-bc] (edge-bc) {\pred{edge}(b,c)};
                    \node[ruleneuron, right color=teal!30!white, right=0.5*\xgap of 2-ary-ab] (edge-ab) {\pred{edge}(a,b)};
                    
                        
                        \draw[edge] (1-ary-a.east) -> node[above=0.1cm] {\textcolor{magenta}{$[1\!\times\!d]$}} (h0-a.west);
                        \draw[edge] (1-ary-b.east) -> (h0-b.west);
                        \draw[edge] (1-ary-c.east) -> (h0-c.west);
                        
                        \draw[edge] (2-ary-bc.east) -> (h0-b.west);
                        \draw[edge] (2-ary-bc.east) -> (h0-c.west);
                        \draw[edge] (2-ary-ab.east) -> (h0-a.west);
                        \draw[edge] (2-ary-ab.east) -> (h0-b.west);
                        
                        \draw[edge] (2-ary-bc.east) -> (edge-bc.west);
                        \draw[edge] (2-ary-ab.east) -> (edge-ab.west);
                    
                    
                    \node[ruleneuron, right color=olive!30!white, right=\xgap of h0-a] (h1-a) {{$h_1$}(a)};
                    \node[ruleneuron, right color=olive!30!white, below=\ygap of h1-a] (h1-b) {{$h_1$}(b)};
                    \node[ruleneuron, right color=olive!30!white, below=\ygap of h1-b] (h1-c) {{$h_1$}(c)};
                    
                        
                        \draw[edge] (h0-a.east) -> node[above=0.1cm] {\textcolor{magenta}{$[d\!\times\!d]$}}  (h1-a.west);
                        \draw[edge] (h0-b.east) -> (h1-b.west);
                        \draw[edge] (h0-c.east) -> (h1-c.west);
                        
                        \draw[edge] (h0-c.east) -> (h1-b.west);
                        \draw[edge] (edge-bc.east) -> (h1-b.west);
                        \draw[edge] (h0-b.east) -> (h1-a.west);
                        \draw[edge] (edge-ab.east) -> (h1-a.west);
                    
                    
                    \node[ruleneuron, right color=olive!30!white, right=\xgap of h1-a] (hi-a) {{$h_{L}$}(a)};
                    \node[ruleneuron, right color=olive!30!white, below=\ygap of hi-a] (hi-b) {{$h_{L}$}(b)};
                    \node[ruleneuron, right color=olive!30!white, below=\ygap of hi-b] (hi-c) {{$h_{L}$}(c)};
                    
                        
                        \draw[edge, dotted] (h1-a.east) -> node[above=0.1cm] {\textcolor{magenta}{$[d\!\times\!d]$}}  (hi-a.west);
                        \draw[edge, dotted] (h1-b.east) -> (hi-b.west);
                        \draw[edge, dotted] (h1-c.east) -> node[below=0.1cm] {$\times$L} (hi-c.west);
                        
                        \draw[edge, dotted] (h1-c.east) -> (hi-b.west);
                        \draw[edge, dotted] (edge-bc.east) -> (hi-b.west);
                        \draw[edge, dotted] (h1-b.east) -> (hi-a.west);
                        \draw[edge, dotted] (edge-ab.east) -> (hi-a.west);

                    
                    \node[ruleneuron, right color=red!30!white, right=\xgap of hi-a] (app-stack-ba) {\pred{app\_stack}(b,a)};
                    \node[ruleneuron, right color=red!30!white, below=\ygap of app-stack-ba] (app-stack-bc) {\pred{app\_stack}(b,c)};
                    
                    \node[ruleneuron, right color=red!30!white, below=2*\ygap of app-stack-bc] (app-putdown-b) {\pred{app\_putdown}(b)};
                    
                        \draw[edge] (hi-a.east) -> node[above=0.1cm] {\textcolor{magenta}{$[d\!\times\!d]$}}  (app-stack-ba.west);
                        \draw[edge] (hi-b.east) -> (app-stack-ba.west);
                        \draw[edge] (hi-b.east) -> (app-stack-bc.west);
                        \draw[edge] (hi-c.east) -> (app-stack-bc.west);
                        
                        \draw[edge] (hi-b.east) -> (app-putdown-b.west);

                    \node[ruleneuron, right color=orange!30!white, below=7*\ygap of app-putdown-b] (wpb-c) {\wpb(c)};
                        
                        \draw[edge] (hi-c.east) -> (wpb-c.west);
                    
                    
                    \node[ruleneuron, right color=blue!30!white, right=1.2\xgap of app-stack-bc] (stack-bc) {\pred{stack}(b,c)};
                    
                        \draw[edge] (app-stack-bc.east) -> node[above=0.1cm] {\textcolor{magenta}{$[d\!\times\!1]$}}  (stack-bc.west);
                        \draw[edge] (wpb-c.east) -> (stack-bc.west);
                    
                    
                    \node[empty, right=0.6*\xgap of stack-bc, label={[blue]above:label}] (y1) {y=1};
                    \draw[dashed, ->] (y1.west) -> (stack-bc.east);
        
                \end{scope}      
            \end{scope} 

        \end{tikzpicture}
	}
	\caption{
		Visualisation of the LRNN architecture. Logical representations of two states (left) form inputs into the Datalog program (middle) that induces differentiable computation graphs (right, partially displayed) for predicting action scores.
	}
	\label{fig:lrnn}
\end{figure*}

%% file: sections/experiments.tex
\newcommand{\baseline}{\textsc{BKPolicy}}
\newcommand{\bk}{\textsc{BK}}
\newcommand{\bwopt}{\textsc{Perfect}}
\newcommand{\scorpion}{\textsc{Scorpion}}
\newcommand{\lama}{\textsc{Lama}}
\newcommand{\lrnn}{\textsc{Lrnn}}
\newcommand{\lrnnSeeds}{3}
\newcommand{\bkSeeds}{3}

\section{Experiments}\label{sec:experiments}
\subsection{Setup}
\subsubsection{Benchmarks}

We perform experiments on 4 classical planning domains: Blocksworld, Ferry, Rover, and Satellite.
We modify Rover to remove the path finding component.
We take the training and test tasks of the domains from the International Planning Competition 2023 Learning Track (IPC23LT)~\cite{seipp.segoviaaguas.2023}.
Other domains from the IPC23LT are omitted as they were either too easy by exhibiting fast optimal algorithms, or too difficult by having no polynomial time satisficing algorithms.

\begin{figure}[t]
    \centering
    \newcommand{\vcnt}[1]{\raisebox{-0.5\height}{#1}}
    \newcommand{\vcntb}[1]{\raisebox{-0.55\height}{#1}}
    \newcommand{\ccc}{0.2}
    \newcommand{\cccb}{0.5}
    \newcommand{\jump}{0.12em}
    \vcntb{\includegraphics[width=0.6\columnwidth]{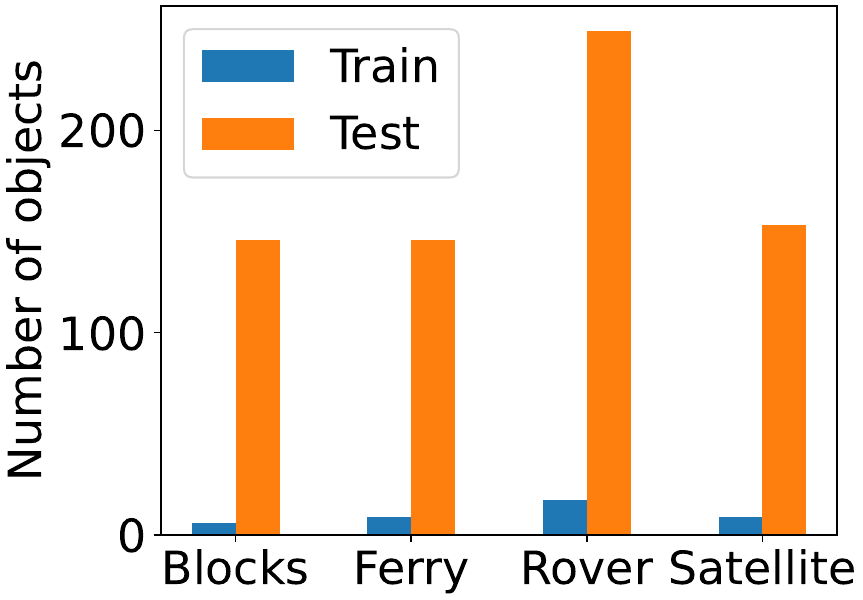}}
    \caption{Range of task sizes in terms of the number of objects in the training and testing tasks.
    }
    \label{fig:objects}
\end{figure}

We generate training data labels by expanding the state space of training tasks with less than 10000 states.
The data has the form of $(s, a, y)$ tuples where $s$ is a state with the goal condition encoded, $a$ is an applicable action in $s$, and $y \in \sett{0, 1}$ indicates whether the action is optimal or not.
At most 25 training tasks are selected for each domain in this way.
The data is used to train the LRNNs to compute a policy from BK that aims to minimise plan length.
The bar plot in the left of Figure~\ref{fig:objects} shows the sizes of train and test tasks in terms of the number of objects, noting that the training tasks are significantly smaller than the testing tasks.

The ground truth we compare against for a task is its optimal plan length.
We compute the optimal plan length with the \bwopt{} solver~\cite{slaney.thiebaux.2001} for Blocksworld, and \scorpion{}~\cite{seipp.etal.2020} for the remaining domains.
Given the difficulty of computing optimal plan lengths, we focus on easy and medium tasks from the IPC23LT and note that \scorpion{} does not generate optimal plans for all tasks in the computational budget.
All training procedures, policy execution, and planner baselines, as will be described, are run entirely on CPUs on a cluster with a time limit of 3600 seconds.

\subsubsection{Baselines}
We consider two baselines.
The first involves running the satisficing policies (\baseline{}) encoded in the BK Datalog programs which can solve tasks in each domain in polynomial time but not optimally.
The execution of the policies is described in the Policy Execution subsection below.
The second involves running the \lama{} planner~\cite{richter.westphal.2010} and taking the first output plan.
We do not focus on its anytime solving aspect as our approach focuses on generating high quality plans quickly rather than optimising with search.

\subsubsection{Training Parameters}
We train a new LRNN for each domain by optimising the cross-entropy loss on the training data with the Adam optimiser~\cite{kingma.ba.2015} for 100 epochs with a fixed learning rate of $10^{-4}$.
No validation set is used, and weights are chosen from the epoch with the highest F1-score on the entire training data.
We experiment with $L \in \sett{1,2}$ message passing layers, $H \in \sett{8, 16}$ hidden dimensions, and a $\mathsf{max}$ aggregation.
We denote $\lrnn_L^H$ as the LRNN model with $L$ layers and a hidden dimension of $H$.
Each LRNN hyperparameter configuration is trained for \lrnnSeeds{} seeded repeats.

\subsubsection{Policy Execution}
The \baseline{} models are executed by repeatedly applying a uniformly sampled action from $\sigma(s)$ in the current state $s$ until the goal is reached.
\baseline{} experiments are run for \bkSeeds{} seeded repeats to account for variance introduced by sampling.
The $\lrnn$ policies are executed in a similar fashion but instead of sampling uniformly from $\sigma(s)$, we take the action with the highest corresponding score computed by the network, breaking ties arbitrarily.

\subsection{Results}
\newcommand{\pli}{\text{PLI}}
\newcommand{\npli}{\text{NPLI}}

\begin{figure*}[t]
    \newcommand{\vcnt}[1]{\raisebox{-0.5\height}{#1}}
    \newcommand{\vcntb}[1]{\raisebox{-0.55\height}{#1}}
    \newcommand{\ccc}{0.145}
    \newcommand{\cccb}{0.5}
    \newcommand{\jump}{0.12em}
    \vcnt{\includegraphics[height=\ccc\textheight]{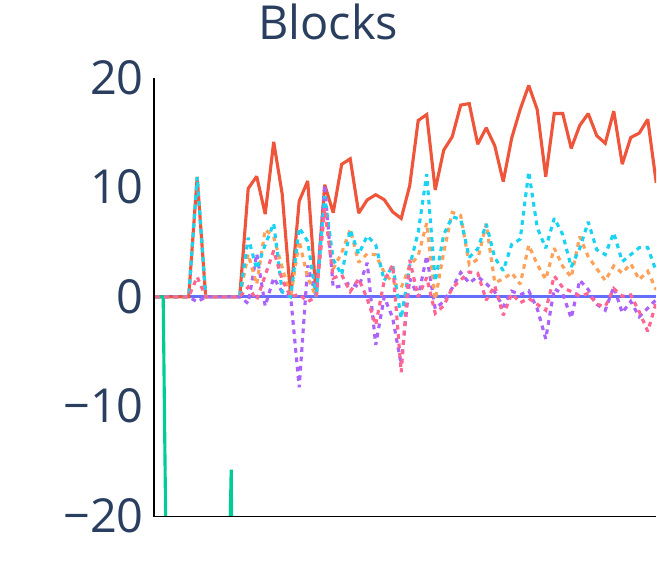}}
    \vcnt{\includegraphics[height=\ccc\textheight]{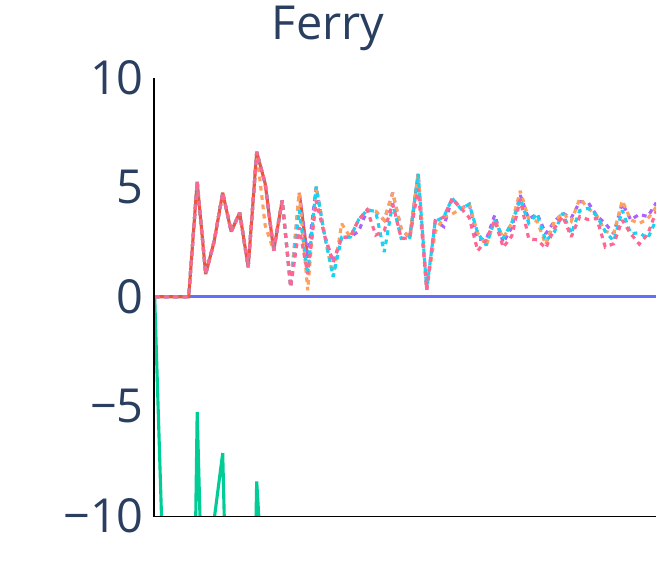}}
    \vcnt{\includegraphics[height=\ccc\textheight]{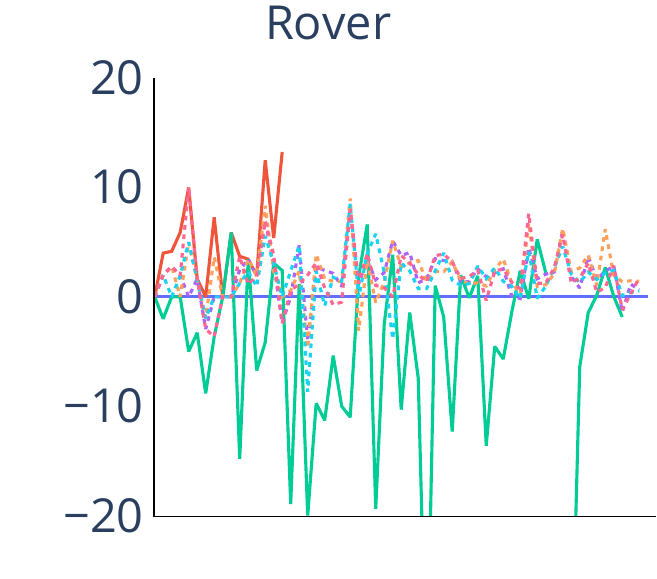}}
    \vcnt{\includegraphics[height=\ccc\textheight]{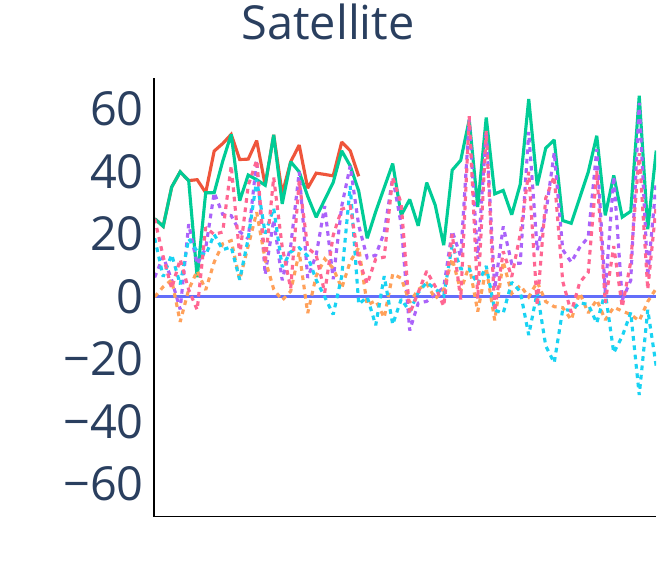}}
    \quad
    \raisebox{-0.45\height}{\includegraphics[height=0.08\textheight]{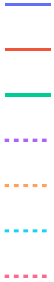}}
    \!\!\!
        {\tiny
        \begin{tabular}{l}
            \baseline{} \\[\jump]
            Optimal \\[\jump]
            \lama{} \\[\jump]
            $\lrnn_1^{8}$ \\[\jump]
            $\lrnn_1^{16}$ \\[\jump]
            $\lrnn_2^{8}$ \\[\jump]
            $\lrnn_2^{16}$ \\[\jump]
        \end{tabular}}
    \caption{Average plan length improvement (\pli{}), computed by Eqn.~\ref{eqn:pli}, over the baseline BK policies ($y$-axis) across tasks of increasing difficulty ($x$-axis). Baselines and planners are denoted with solid lines, and LRNN models with dotted lines.}
    \label{fig:graphs}
\end{figure*}

\newcommand{\pliInput}{x}
\newcommand{\pliInputOpt}{x^*}
\newcommand{\avg}[1]{\braket{#1}}
\newcommand{\bkx}{\bk{}_{\problem}}

Figure~\ref{fig:graphs} displays the average plan length improvement (\pli{}) of the \lrnn{} models over the baseline BK policies for each task in a domain, indicated by the dotted lines, with solid red lines representing the upper bound of improvement and green lines representing the performance of \lama{}.
These scores are computed by
\begin{smallAlign}
    \pli_\problem(\pliInput) = 
    100 \cdot 
    (\avg{\bkx} - \pliInput) / \avg{\bkx}
    \label{eqn:pli}
\end{smallAlign}
where $\pliInput$ is the input plan length, $\problem$ is the task, and $\avg{\bkx}$ denotes the average plan length of the baseline \baseline{} across all seeds on $\problem$.
Table~\ref{tab:improvement_scores} quantifies the graphs in Fig.~\ref{fig:graphs}.
The \pli{} scores are first normalised by dividing by the \pli{} of the optimal plan length, given by
\begin{smallAlign}
    \npli_\problem(\pliInput) = 100 \cdot 
    \pli_\problem(\pliInput) / \pli_\problem(\pliInputOpt) 
    \label{eqn:npli}
\end{smallAlign}
where $\pliInputOpt$ is the optimal plan length for $\problem$.
The normalised scores are then averaged over all tasks where the optimal plan length was computed.
We analyse our experimental results by answering the following questions.

\begin{table}[t]
    \include{tables/improvement_scores.tex}
    \caption{Average and standard deviation of normalised plan length improvement (\npli{}), computed by Eqn.~\ref{eqn:npli}, of LAMA and LRNN models over baseline BK policies for each domain. Higher scores are better, and are capped at 100.}
    \label{tab:improvement_scores}
\end{table}

\subsubsection{Does learning improve over BK policy plan quality?}
From Tab.~\ref{tab:improvement_scores}, the \lrnn{} policies on average outperform the baseline BK policies in terms of plan quality across all domains, with the best \lrnn{} configuration for each domain achieving over 50\% of the possible plan length improvement with respect to the optimal plan length.
From Fig.~\ref{fig:graphs}, the \lrnn{} policies also generally outperform the baseline on tasks where we cannot compute the optimal plan length.
Two exceptions are $\lrnn_2^8$ and $\lrnn_1^{16}$ for Satellite which decrease in performance on larger tasks.
We further note that both the \baseline{} and \lrnn{} policies return significantly shorter plans than \lama{} on all domains except Satellite.

\subsubsection{Which BK properties are important for plan quality?}
Property~\ref{ass:deadend} ensures that both the baseline \baseline{} and \lrnn{} policies never fail and eventually achieve the goal, and is satisfied by all encoded policies.
Assuming Property~\ref{ass:deadend} is satisfied, we discover that Property~\ref{ass:cycle} is more important than Property~\ref{ass:optimal} for achieving lower plan lengths.
More specifically, the Satellite BK policy in Example~\ref{eg:satellite} does not preserve optimal actions in 0.11\% of training states due to encoding the turn\_to action priorities.
Removing such priorities preserves optimal actions and hence Property~\ref{ass:deadend} but introduces cycles in the form of being able to take arbitrary turn\_to actions.
Informal experiments show that allowing cycles in such BK policies results in significantly longer plans for both the \baseline{} and \lrnn{} policies.
This is because cycles in a plan result in sequences of redundant actions due to states being history-independent.
Although such actions can be removed with plan postprocessing techniques~\cite{bercher.etal.2024}, the policy execution is inefficient due to revisiting previously seen states.

\subsubsection{What are the effect of hyperparameters?}
Results displayed in Tab.~\ref{tab:improvement_scores} suggest no statistically significant conclusion regarding the effect of hyperparameters.
Each domain has a different hyperparameter configuration that performs best, and there is no clear relationship between increasing the hidden dimension size $H$ or message passing layers $L$ and performance.
We note however that models with smaller $H$ and $L$ are significantly faster to train and execute.

\subsubsection{How long does training take?}
We note that generating the training data from expanding state spaces of small tasks take less than $5$ seconds for each domain.
The \lrnn{} model training ranges between $156$ to $3522$ seconds depending on the domain and hyperparameter configurations on CPUs, and could be significantly sped up with access to GPUs.
In other words, training is rather cheap and the cost of training can be amortised when solving tasks within the domain, as \lrnn{} policies have polynomial time execution for polynomial domains, while domain-independent planners such as \lama{} have worst case exponential time complexity.

%% file: tables/improvement_scores.tex

\newcommand{\cellMacro}[2]{\text{\scriptsize{#1}} \text{\tiny$\tiny{\;\!\pm\;\!}$} \text{\tiny{#2}}}
\newcommand{\domainCell}[1]{{#1}}\small
\begin{tabularx}{\columnwidth}{X r r r r}
    \toprule
    Planner & \domainCell{Blocks} & \domainCell{Ferry} & \domainCell{Rover} & \domainCell{Satellite} \\
    

    \midrule
    \lama & $\cellMacro{-674.4}{603.5}$ & $\cellMacro{-504.5}{659.8}$ & $\cellMacro{-28.4}{147.8}$ & $\cellMacro{88.4}{17.2}$ \\
    $\lrnn_{1}^{8}$ & $\cellMacro{46.4}{34.1}$ & $\cellMacro{94.0}{20.0}$ & $\cellMacro{57.4}{36.7}$ & $\cellMacro{16.8}{16.8}$ \\
    $\lrnn_{1}^{16}$ & $\cellMacro{22.2}{49.5}$ & $\cellMacro{100.0}{0.0}$ & $\cellMacro{56.3}{40.2}$ & $\cellMacro{50.8}{27.8}$ \\
    $\lrnn_{2}^{8}$ & $\cellMacro{25.8}{45.2}$ & $\cellMacro{95.7}{11.7}$ & $\cellMacro{53.1}{48.2}$ & $\cellMacro{50.5}{30.3}$ \\
    $\lrnn_{2}^{16}$ & $\cellMacro{53.2}{31.6}$ & $\cellMacro{97.1}{9.9}$ & $\cellMacro{44.9}{40.2}$ & $\cellMacro{28.7}{26.2}$ \\
    \bottomrule
\end{tabularx}

%% file: sections/related-work.tex
\section{Related Work}\label{sec:related-work}
Our work represents one of the many emerging research directions in scaling up planning, moving beyond the traditional paradigm of directly inputting a task into a domain-independent planner and hoping it solves the task.
This section outlines related work concerning different paradigms for scaling up planning: learning for planning, generalised planning, and planning incorporating domain knowledge.
We note that our work spans all three of these areas.

\subsubsection{Learning for Planning}
Learning for Planning is attracting the most attention recently due to the success of ML across various other research fields.
Recent deep learning works involve learning action policies~\cite{toyer.etal.2020,silver.etal.2024,rossetti.etal.2024}, heuristics for guiding search or greedy policies~\cite{shen.etal.2020,karia.srivastava.2021,staahlberg.etal.2022,staahlberg.etal.2023,chen.etal.2024,agostinelli.etal.2024}, and quantifying expressiveness of architectures~\cite{horcik.sir.2024}.
Traditional symbolic or classical machine learning have also been applied to learn more efficient and explainable policies~\cite{frances.etal.2019,hofmann.geffner.2024}, heuristics~\cite{chen.etal.2024a}, and subgoals~\cite{drexler.etal.2024}.
We refer to the survey by~\citet{jimenez.etal.2012} for earlier works on learning for planning.

\citet{khardon.1999} proposed learning decision lists, a subset of Datalog consisting of an ordered list of first-order Horn clauses.
\citet{gretton.thiebaux.2004} learned decision lists entirely from scratch with a new learning algorithm for representing \emph{optimal} general policies and value functions for relational MDPs, with the tradeoff that learned value functions cannot generalise beyond the range of values seen in the training data.
Various differentiable inductive logic programming techniques have also been explored for planning~\cite{dong.etal.2019} and Reinforcement Learning settings~\cite{hazra.raedt.2023}.
Orthogonally, researchers have studied different optimisation criteria better suited for learning in planning contexts~\cite{garrett.etal.2016,orseau.etal.2023,chrestien.etal.2023,hao.etal.2024}.

\subsubsection{Generalised Planning}
Generalised Planning (GP) entails computing programs that characterise the solutions of planning tasks in a domain~\cite{srivastava.etal.2008}.
Policies, as described in this work, constitute one such program. 
Other characterisations include finite state controllers~\cite{bonet.etal.2009,bonet.etal.2010,hu.giacomo.2011,hu.giacomo.2013,aguas.etal.2018} and programs with branching and loops~\cite{aguas.etal.2021,aguas.etal.2022}.
GP has also been represented as Qualitative Numeric Planning (QNP) tasks~\cite{srivastava.etal.2008} and has been shown to be theoretically equivalent to fully observable nondeterministic planning (FOND)~\cite{bonet.geffner.2020}.
The connection between GP and FOND has also been exploited to synthesise generalised policies~\cite{bonet.geffner.2018,illanes.mcilraith.2019}.
For comprehensive surveys on GP, we refer to articles by~\citet{celorrio.etal.2019} and~\citet{srivastava.2023}.

\subsubsection{Planning Incorporating Domain Knowledge}
Incorporating domain knowledge into planning primarily involves deciding the language to formally represent such knowledge.
%
Hierarchical Task Networks~\cite{bercher.etal.2019} in the SHOP planner~\cite{nau.etal.1999,nau.etal.2003}, and temporal logics in TLPlan~\cite{bacchus.kabanza.2000} have been used to guide search.
\citet{baier.etal.2008} compile domain knowledge represented in a Golog-inspired language into additional planning predicates and conditional actions.
%
The PDDL language can also be extended with axioms~\cite{thiebaux.etal.2005} which can be used to encode the provided Datalog background knowledge or by restricting the structure of planning tasks in a domain~\cite{grundke.etal.2024}.
Reward Machines~\cite{icarte.etal.2022} allow for expressive reward function modelling via finite state models to improve the encoding of RL domains and performance of agents.
Domain-Independent Dynamic Programming~\cite{kuroiwa.beck.2023,kuroiwa.beck.2024} is a declarative problem solving language inspired by PDDL for combinatorial optimisation which allows user input to guide the solving process and yields competitive performance compared to MIP and CP solvers.
%